\documentclass{bmvc2k}

\title{Generative Cooperative Net for\\ Image Generation and Data Augmentation}

\addauthor{Qiangeng Xu}{qiangeng@buaa.edu.cn}{1}
\addauthor{Zengchang Qin}{zcqin@buaa.edu.cn}{1}
\addauthor{Tao Wan}{wantao@buaa.edu.cn}{1}

\addinstitution{
 Intelligent Computing and Machine Learning Lab\\
 Beihang University \\
 Beijing, China
}

\runninghead{Xu, Qin and Wan}{GENERATIVE COOPERATIVE NETWORK}

\def\etal{\emph{et al}\bmvaOneDot}

\begin{document}

\maketitle

\begin{abstract}
How to build a good model for image generation given an abstract concept is a fundamental problem in computer vision.
In this paper, we explore a generative model for the task of generating unseen images with desired features. We propose the Generative Cooperative Net (GCN) for image generation. The idea is similar to generative adversarial networks except that the 
generators and discriminators are trained to work accordingly. 
Our experiments on hand-written digit generation and facial expression generation show that GCN's two cooperative 
counterparts (the generator and the classifier) can work together nicely and achieve promising results. We also discovered a usage of such generative model as an data-augmentation tool. Our experiment of applying this method on a recognition task shows that it is very effective comparing to other existing methods. It is easy to set up and could help generate a very large synthesized dataset.
\end{abstract}

%-------------------------------------------------------------------------
\section{Introduction}
\label{sec:intro}

\begin{figure}
\begin{center}
\includegraphics[width= 9.8cm]{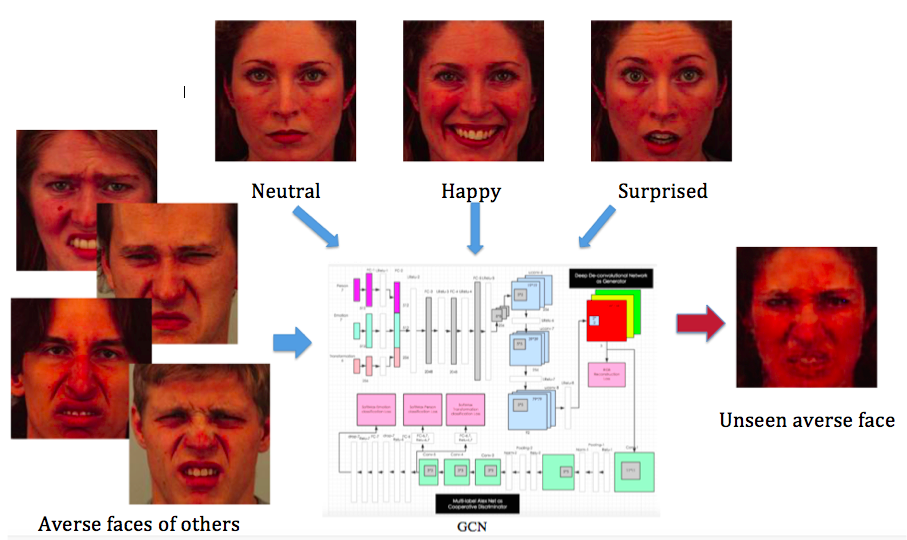}
\end{center}
\caption{ An example of learning high-level concepts such as facial expression using GCN. The model takes faces of (a) Neutrality, (b) Happiness (c) Surprise of a person as well as facial expressions (including aversion) from other people. 
We can generate an aversion face of this person by learning the abstract concept of aversion. }
\label{fig:teaser}
\end{figure}

Generative network for image generation has been an active research area and have been applied to various computer vision applications such as  super-resolution \cite{ledig2016photo}, image painting \cite{pathakCVPR16context}, manifold learning \cite{zhu2016generative} and semantic segmentation \cite{Noh_2015_ICCV}. A wide variety of deep learning approaches involve generative parametric models. Many models are with encoder-decoder 
structure that can reconstruct the image from the latent representation. 
Different from those previous works, this research focus on learning the high-level concepts and generating unseen images with desired features (e.g. the task shown in Figure \ref{fig:teaser}). To solve this sort of tasks, we developed a new generative network model, the \emph{Generative Cooperative Net} (GCN).

In our research, we also observed that a well-trained GCN could provide an highly efficient alternative for data-augmentation, other than those traditional transformation methods which could only provide less complex variation.
In this paper, our major contribution is twofold:
%\begin{itemize}
  % \item 
 (1) We proposed a new generative model, Generative Cooperative Net (GCN) to learn the high-level concepts of images' latent description and generate large($158 \times 158$) high-quality unseen images with desired features. 
(2) We discovered a way to use GCN as an effective data-augmentation tool for recognition tasks. 
%\end{itemize}

\section{Related Work}
\label{sec:relatedwork}

%\subsection{Image generation}

Although the generative model for image generation has been well-studied by using methods such as "Deep Boltzman Machine" \cite{salakhutdinov2009deep}, the quality and capability of the generative model was boosted especially after researchers started to "reverse" the top-down deep learning network with up-scale structures such as "de-convolutional layer" introduced by Zeiler \etal~\cite{Zeiler01}.
One type of the generative networks (e.g. the model introduced in \cite{Dosovitskiy01}) are parametric and applied supervised learning to project the high level description to the corresponding images. Although such structure can build perfect projection for the label-image pairs, those networks always suffer from "memorizing" the training set and have limited ability on knowledge transfer. 

Another family of deep generative networks adopted the Generative Adversarial Networks (GANs) introduced by Goodfellow \etal~\cite{Goodfellow01}. Such frame work provides an attractive alternative to supervised learning methods and enable the networks to learn the data distribution of the objects' manifold. Certain models can generate realistic images and be very creative, meanwhile providing limited controllability over the desired features. To extend the controllability, Mirza \etal~\cite{mirza2014conditional} introduced the "Conditional GANs" which learns a conditional distribution that would guide the model to generate desired features. Yet, the network still need a large dataset to learn the manifold of the concept. 
Our research has adopted the supervised learning methods to learn the concept directly. Since we don't need to learn the distribution of a visual concept compared to the models using GANs, the dataset we need is much smaller. On the other hand, the classifier in our framework could facilitate the knowledge transformation, which makes our model a good choice among the supervised generative models. 

%\subsection{Data Augmentation in Object Recognition}

Object recognition has become a mature research area since the fast development of deep convolution networks\cite{Alex01,Simonyan14c,Szegedy_2015_CVPR,He_2016_CVPR}. To prevent overfitting and data starvation, all of those studies have adopted data augmentation methods such as random cropping, scale variation and affine transformation. Although those methods are very effective to reduce the impact of images' details, recognition of high-level features such as identity and facial expression still require a sufficient amount of images of the target. We also studied some face augmentation methods including landmark perturbation and synthesis methods on hairstyles, glasses, poses and illuminations \cite{lv2017data}. Those methods although could provide high-level augmentation, most of them rely on the specific 3D knowledge and well-studied human-face model. Instead, our GCN's data-augmentation method is much easier to be set up and applied to many other tasks.

\section{Method and Model Architecture}

\begin{figure}
\begin{center}
\includegraphics[width= 11cm]{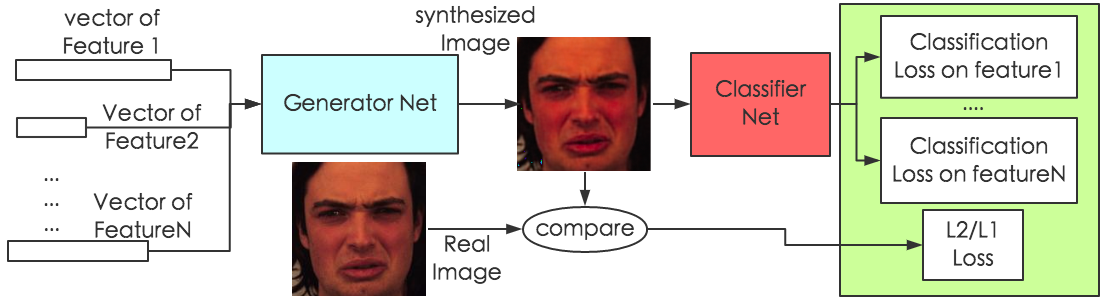}
\end{center}
\caption{The loss inside the green box will be back-propagated to the classifier and the generator accordingly. }
\label{fig:teaser1}
\end{figure}

Our goal is to obtain an unseen image with desired features. To better achieve this goal, we designed the model with two components, a generator and a classifier. The generator we use mainly adopt the architecture from the network introduced by Dosovitskiy \etal~\cite{Dosovitskiy01} and the classifier is based on the classic AlexNet \cite{Alex01}. 
The form of the input is a set of high-level feature vectors and their corresponding images. \[D = \{(f_1^1,f_1^2,\dotsc,f_1^M),(f_2^1,f_2^2,\dotsc,f_2^M),\dotsc,(f_N^1,f_N^2,\dotsc,f_N^M)\}\] where $M$ is the amount of features each image has and $N$ denotes the amount of images we have in the training set. In terms of the facial expression generation task, the features we used in training are: people's identity (a 70 element one-hot encoding vector), the category of his or her expression and the transformation being applied.
% \( = \{(P_1,E_1,T_1),(P_2,E_2,T_2),\dotsc,(P_N,E_N,T_N)\}\) 
We used \(\{I_r^1,I_r^2,\dotsc,I_r^N\}\) to denote the real images in the training set and \(\{I_s^1,I_s^2,\dotsc,I_s^N\}\) as the synthesized images. The whole training process is described in Figure \ref{fig:teaser1}.

\subsection{Objective Function}
Since each image has multiple features, we need to test on each of them and adjust the weights to get an all-rounded result (e.g. a smiling face of person No.34 should look like person No.34 and can be recognized as smiling). Instead of using multiple classifier, we only used one multi-labeled classification network, which produced cross entropy loss on each feature.
The objective of the classifier is: 
\begin{equation}
\min_W - \frac{1}{N} \left[ \sum_{i=1}^{N} \sum_{f=1}^{M} \sum_{j=1}^{K_f}  Weight_f \left\{y_f^{(i)} = j\right\} \log \frac{e^{\theta_{f_j}^T x_f^{(i)}}}{\sum_{l=1}^{K_f} e^{ \theta_{f_l}^T x_f^{(i)} }}\right]
\end{equation}
In which $N$ is the amount of images in a running batch, $M$ is the amount of features for one sample and \(K_f\) is the class dimension for a specific feature $f$. Each classification loss has a weight to balance the dimension.
According to the research in \cite{Dosovitskiy01} and \cite{Isola0}, the pixel to pixel loss is essential to the detail generation. Besides, without the pixel to pixel loss, we found it difficult for generator to create sensible results or even move towards right direction in initial batches. Therefore, the objective of the generator has two components: (1) the pixel to pixel Euclidean loss while compared with the real image, and (2) the classification loss created by the classifier.
{\small
\begin{multline}
 \min_W  \frac{1}{N} \{ \sum_{i=1}^{N}||Pixel_r^i -  Pixel_s(f_{i_1},...,f_{i_M})^i ||^2  -  \left[ \sum_{i=1}^{N} \sum_{f=1}^{M} \sum_{j=1}^{K_f}  Weight_f \left\{y_f^{(i)} = j\right\} \log \frac{e^{\theta_{f_j}^T x_f^{(i)}}}{\sum_{l=1}^{K_f} e^{ \theta_{f_l}^T x_f^{(i)} }}\right] \} 
\end{multline}
}

\begin{figure}
\begin{center}
\includegraphics[width= 12cm]{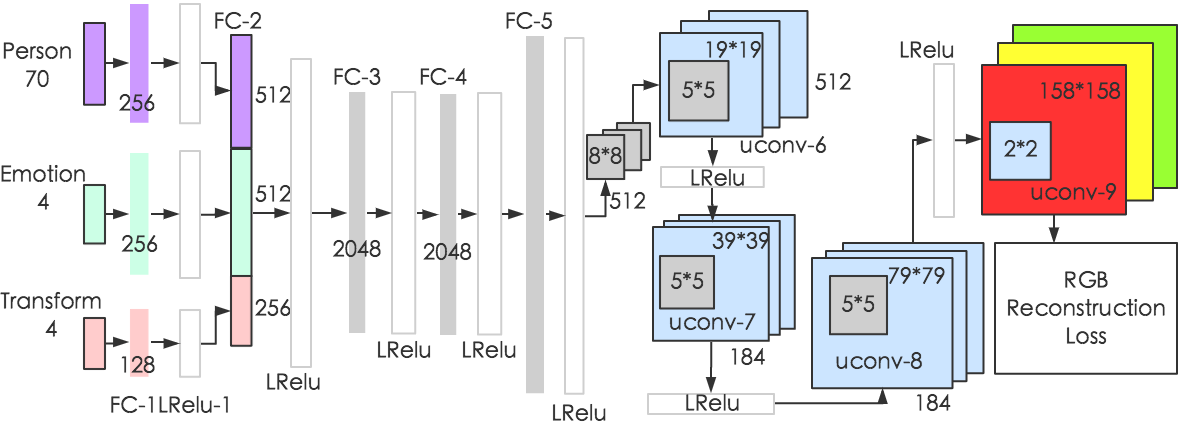}
\end{center}
\caption{The architecture of the generator in GCN}
\label{fig:gen}
\end{figure}
\subsection{Architectures}

We have studied the generation networks introduced in \cite{Dosovitskiy01}, \cite{DBLP:journals/corr/RadfordMC15} and \cite{Noh_2015_ICCV}. We first of all used two fully connected layers for each feature to extent up to 256 dimensions, then concatenated those layers of each individual feature and applied another two fc-layers on them. We then, reshaped the vector of the fifth fc-layer to a $8 \times 8$ matrix with 256 channels. From here, we applied fractional-strided convolutions to replace up-pooling strategy used in most generative models. Each deconvolutional layer's stride has been set to 2. According to the output size relationship: \( size_{out} = (size_{in} - 1) \times stride + kernel_size - 2 \times pad \) we modified each layers to meet the target dimension of the final output layers (in case of face expression generation which is \(158 \times 158\), we have 4 such layers). The details of the generator's structure is shown in Figure \ref{fig:gen}. We also tested different types of activation layers and found out the LeakyReLU with negative slop of 0.1 could outperform RelU, PRelU and other commonly-used activation layers.

\begin{figure}
\begin{center}
\includegraphics[width= 12cm]{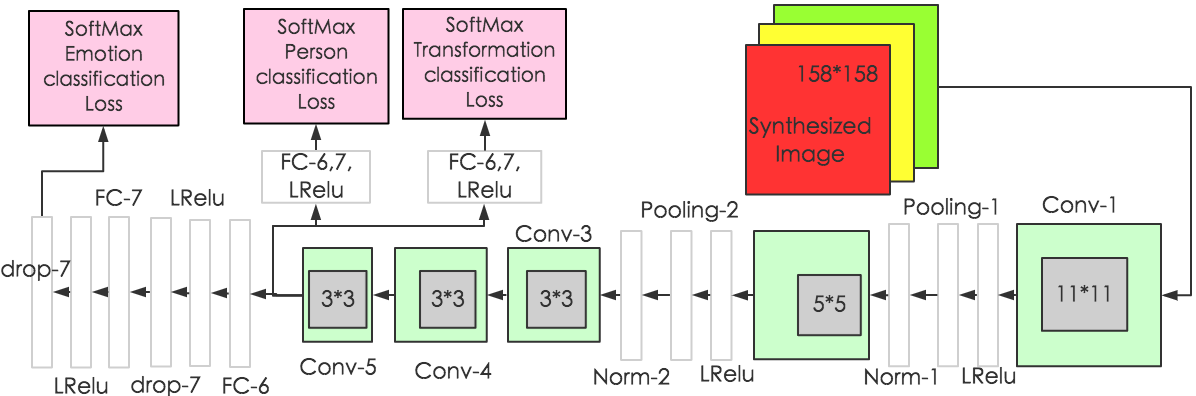}
\end{center}
\caption{Structure of the classifier: a multi-task AlexNet }
\label{fig:class}
\end{figure}

%\subsection{Classifier architecture}
The classifier's structure is based on the AlexNet introduced by Alex \etal~\cite{Alex01}. We could in fact use more complicated deep network such as VGG \cite{Simonyan14c} or even the 152 layer ResNet introduced by He \etal~\cite{He_2016_CVPR}. However, after considering the memory usage, training speed and most importantly, the gradient degradation problem in deep learning (17 layers of the generator plus layers of the classifier), we decided to stick with AlexNet. Different from the original version, we assigned two fc-layers and a softmax layer for each feature separately, which makes it a multi-task classifier. After exploring the activation layers, we found out LeakyReLu layers with negative slop of 0.2 would outperform the original ReLu layers. Besides, we also reduced the stride of first convolutional layer since our image resolution is smaller than the ImageNet. The structure of our classifier is shown in Figure \ref{fig:class}.

\subsection{Image Augmentation}

We conducted a data-augmentation experiment by using a trained GCN. Since we observed that the GCN is also capable to synthesize two people's face, we generated all combination of two people's face under every emotion. We then used them to form a much larger dataset and performed emotion recognition training on it. In the end, we compared the accuracy achieved on the synthesized dataset with the accuracy trained on the original dataset.

\section{Experimental Studies}
\subsection{Dataset}

We here conducted two experiments to explore the task of generating unseen images with desired features. In the first experiment, we used "The Karolinska Directed Emotional Faces (KDEF)" \cite{lundqvist1998karolinska}, 
which is a set of 4900 pictures of human facial expressions of emotion. The dataset contains 70 individuals, each displaying 7 different emotional expressions and each expression being photographed (twice) from 5 different angles. To simplify the generative task, we only selected the front views and picked 4 emotions: neutrality, aversion, happiness and surprise. We used OpenCV's default frontal face detection cascade script to further crop the image and re-sized them to the resolution of \(158 \times 158 \). We also rotated each image by 0, 90, 180 and 270 degree and conducted mirror operation to each degree, which result in 7 different transformations from every original image. 

As for the second generation experiment, we used the MNIST dataset \cite{lecun1998mnist} to train and generate handwriting digits. We select 100 different examples for each digit, and coloured each grey levels image to red, green and blue versions. Same as the first experiment, we also augmented the dataset by rotating each image by 0, 90, 180 and 270 degrees.

\begin{figure}[h]
\begin{center}
\begin{tabular}{ccc}
\bmvaHangBox{\parbox{8.4cm}{~\\
\rule{0pt}{1ex}\hspace{1.0mm}\includegraphics[width=8.4cm]{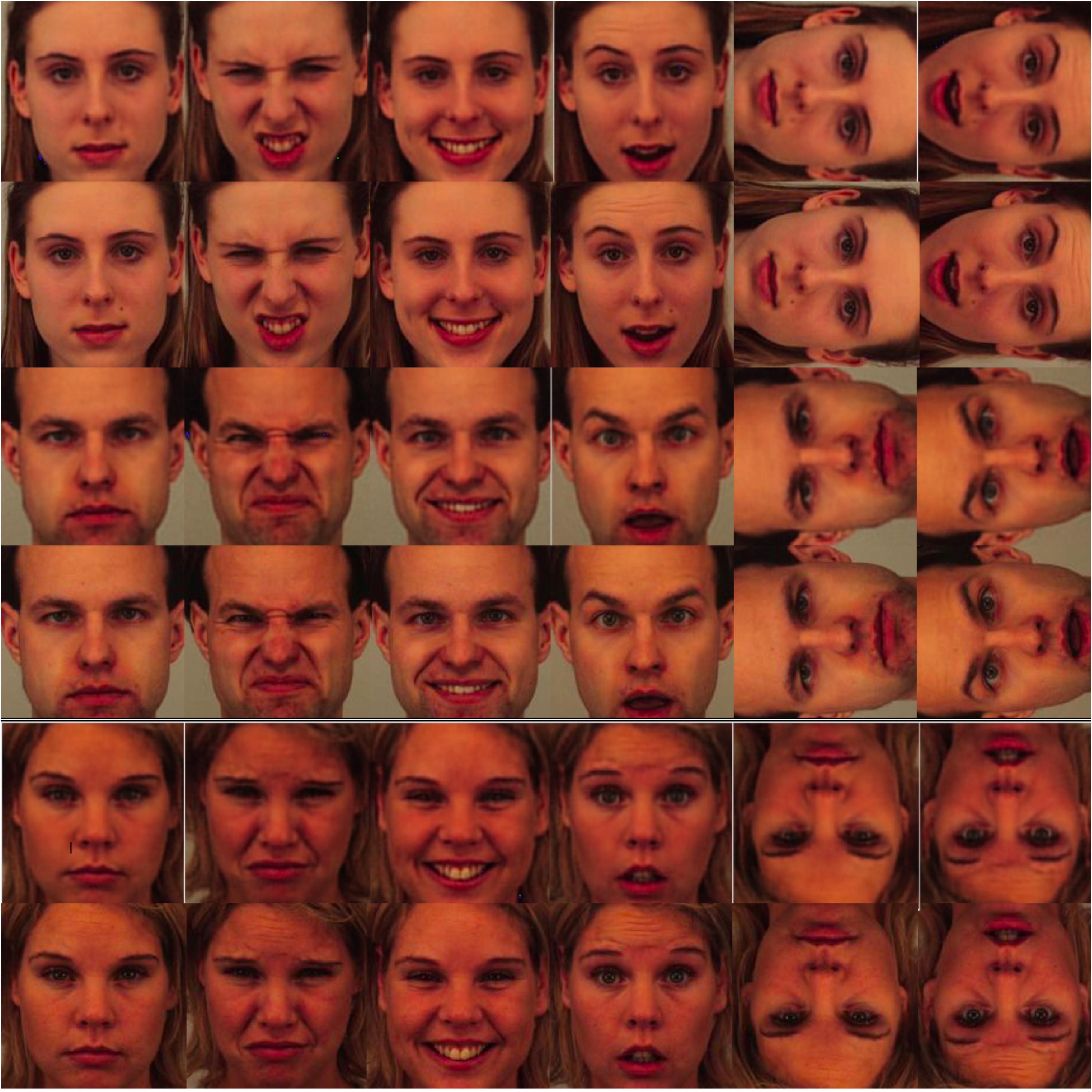}}}&
\bmvaHangBox{\includegraphics[width=3.5cm]{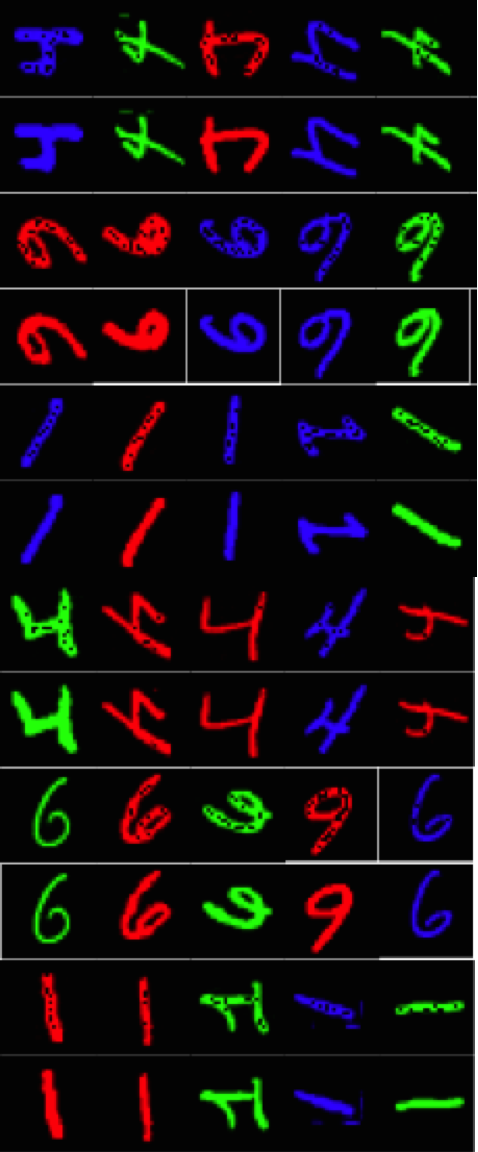}}\\
(a)&(b)
\end{tabular}
\end{center}
\caption{Existing objects reconstruction: (a) The 1st, 3rd and 5th rows are the synthesized faces and the 2nd, 4th and 6th are the real faces in the training set. (b) The odd rows are the synthesized digits and the even rows are the corresponding real digits}
\label{fig:5ab}
\end{figure}

\subsection{Model Training}

%\subsubsection{Image Generation}
Both of our generation experiments are implemented in Caffe \cite{jia2014caffe}. We used Adam \cite{kingma2014adam} with base learning rate of 0.0002. For the face expression generation, we found the momentum \(\beta_1\) as 0.9, \(\beta_2\) as 0.99 and \(\epsilon=10^{-8}\) would make the training most stable. Meanwhile, the digit generation task seems to favor momentum \(\beta_2\) at 0.995. Both tasks have batch size of 64 and the learning rates are divided by 2 after every 1000 batches. We trained both tasks up to 40000 batches and found out the loss have become very stable. Since the handwriting digit is only \(28 \times 28\), we reduced the deconvolutional layers in generator and the convolutional layers in classifier to fit the dimension. 

To test the capability of GCN on generating unseen images, we deliberately selected 10 people that would either miss several transformations or miss 1 facial expression. Then, we tested the trained generator by inputting the unseen expression for those people and evalutated the generated image. As for the MNIST dataset, we only picked out several transformations or colors from a handwriting digit to form our test set, since there is no other high-level features like emotion for this task.
We found setting the loss of emotion and identity to be 10 and euclidean loss to be 1 could produce the best result because of the better balance achieved between perception and details. We also explore other weights combination such as $100$ for the classification and 1 for the Euclidean loss, which made both the generator and classifier hard to converge. 
%This result is caused by poor feed back from classifier and poor input samples provided by generator. To rule out the effect of the bad initial feed back from the classifier, we also explored the strategy of pre-training the classifier with real images. To prevent the noise caused by miss classification of the classifier (3\% error rate in our case),  we also tried on replacing the original softmax loss by the miss-matched softmax loss, which is the euclidean difference between classification result vector of the real image and the result vector of the synthesized image. 

%\subsubsection{Image Augmentation}
To explored GCN's capability of data-augmentation, we trained a GCN by selecting 65 people with one session for all 5 emotions. Then we deployed the well-trained GCN and input 0.5 on each person in every combination of two people. We then, got \(65 \times 65 \times 5\) synthesized images and filtered out images with low quality. We used a single-task emotion classifier to train on the synthesized and the original dataset. The classifier has the same structure of the GCN's classifier. We used Adam with base learning rate of 0.0002, momentum \(\beta_1\) as 0.9 and \(\beta_2\) as 0.999. Both datasets have been trained for 40000 batches with batch size of 64 and the weight decay of 0.0005. The test set included the faces of these people in another session and the faces of other people under these 5 emotions. 

\subsection{Result Analysis}
%\subsubsection{Existing objects reconstruction}
As for the face generation task, if a target feature combination is existed in the training set, the reconstruction result looks very promising. The synthesized images are almost identical to the real images (Figure \ref{fig:5ab}-(a)) and would only miss out some minor details. For the handwriting figure generation task, the reconstructions are fairly good as well, except for some noisy pixels. (Figure \ref{fig:5ab}-(b)).

\subsubsection{Low level feature generation}
\begin{figure}[h]
\begin{center}
\begin{tabular}{ccc}
\bmvaHangBox{\parbox{4.2cm}{~\\
\rule{0pt}{1ex}\hspace{1.0mm}\includegraphics[width=3.8cm]{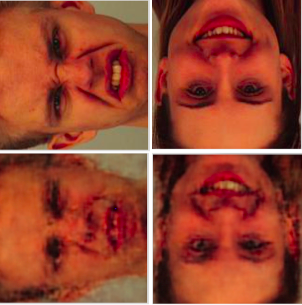}}}&
\bmvaHangBox{\includegraphics[width=6cm]{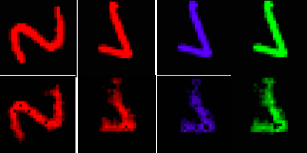}}\\
(a)&(b)
\end{tabular}
\end{center}
\caption{Low level features generation: (a) Images on the first row are ground truth images with 90 degree and 180 degree rotation, which have been taken out of the training set. Images on the second row are the synthesized images. (b) The first column shows the generation result of a 90 degree rotated red digit, when the training set only miss the red color. The other three columns show the generation of a 180 degree rotated "7" when all colors has been taken out for this rotation. }
\label{fig:6ab}
\end{figure}

The low level features we want to generate include the rotational transformation for KDEF and both rotational and coloring transformation on MNIST. For facial expression generation, we took the 90 degree rotation of a person's averse face out of the training set. We then, fast-forward with the input vector of this person's identity, the averse facial expression and the 90 degree rotation. Even the network has never seen the image, our results show that the concept of rotation could be learnt through training (Figure \ref{fig:6ab}-(a)). 
In the digit generation experiment, we first only picked out the red color of a figure "2" and tried to generate the red "2" after training. We also took out all three colors of a digit "7" to test whether GCN can generate the image while 2 features are both missing. As we can see in Figure \ref{fig:6ab}-(b), the results on the "missing color" is better than result on the "missing transformation". It is possible that mathematically, the rotation transformation relationship has more complexity than the color concept, thus more difficult for GCN to learn the relations.

\subsubsection{High level feature generation}
\begin{figure}
\begin{center}
\includegraphics[width= 11cm]{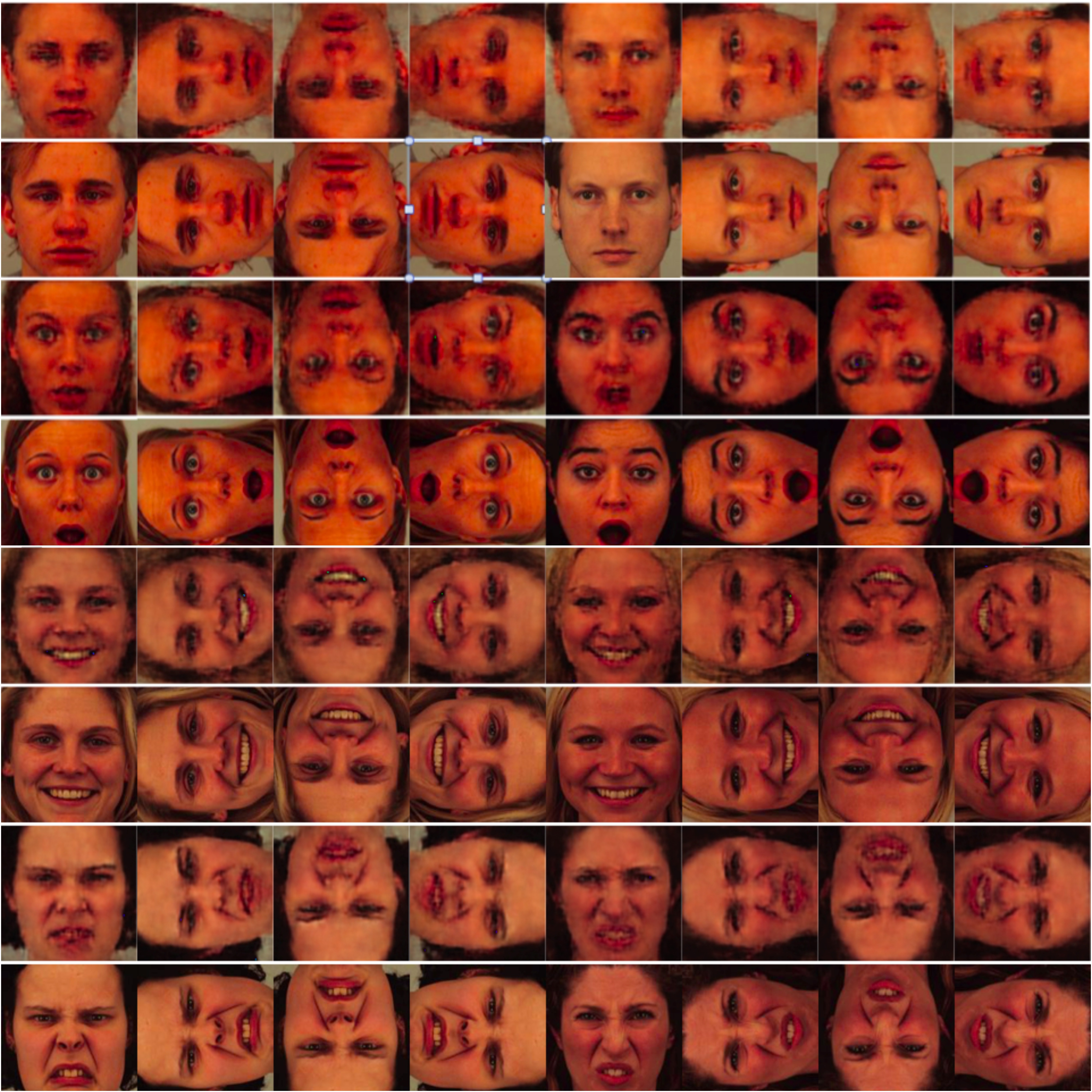}
\end{center}
\caption{High level features generation: the 1st, 3rd, 5th and 7th rows are the synthesized images and the 2nd, 4th, 6th and 8th rows are their corresponding real images. The first two rows show the neutral faces, while the second two rows show the surprised faces, the third two rows are the happy faces and the last two rows are the averse faces.}
\label{fig:highlevel}
\end{figure}

The reason we chose an expression generation task for our research is because human facial expression of emotion could provide two "easy to recognize" high-level features, the emotion and the face identity. Here we select 2 people to test each emotion. We deliberately took out the original images and transformations of this facial expression of the 2 people and tested the network's ability to generate this unseen expression for them. Inspecting the result shown in Figure \ref{fig:highlevel}, we can see the emotions such as happiness and aversion could be nicely learnt and the synthesized images are very close to the ground truth. The emotion such as surprise is a bit harder to train, since in KDEF, the facial expressions of different people on surprise have larger variance (the degree of the mouth expansion and raising eye brows are very different among individuals). More importantly, the person's identity could be well preserved which significantly increased the credibility of the synthesized image. 

\subsubsection{Image Augmentation}
\begin{figure}
\begin{center}
\includegraphics[width= 10.8cm]{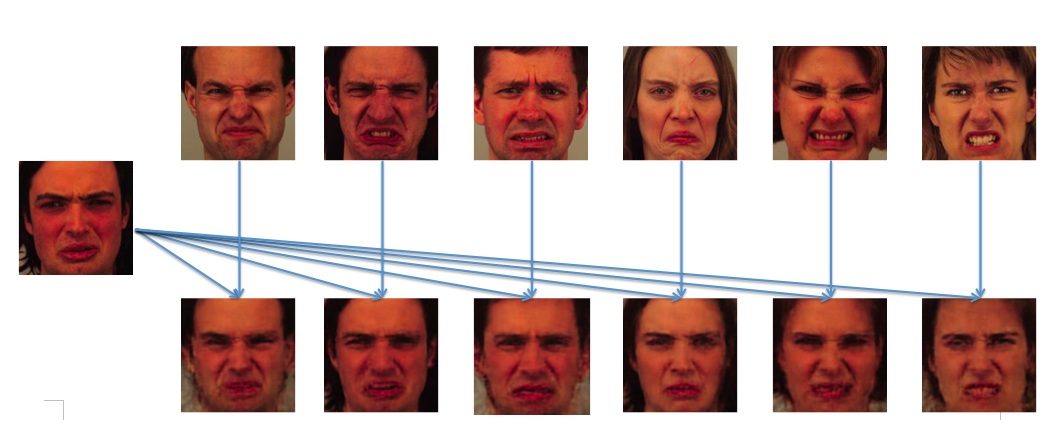}
\end{center}
\caption{An illustration of generating 6 synthesized faces by combining one averse face of one person with averse faces of 6 different people. The synthesized faces can also be used in other recognition tasks such as emotion recognition. }
\label{fig:merge}
\end{figure}

As we have seen from previous results, the "unseen emotion" can be well captured by our new model. The faces  we generated using a combination of every two individual could have a fairly good quality, and can be used
as a training set for recognition tasks. 
Almost every synthesized image can preserve the emotion concept of its parent images and look like both people (Figure \ref{fig:merge} shows an example of averse face generation). By combining people with each others at the ratio of 0.5 to 0.5, we can finally obtain 21125 faces from the original 325 faces. After training with the same hyper-parameter settings, the emotion classification accuracy can boost from 92\% on the original dataset, to 94\% on the synthesized dataset. 

\section{Conclusions and Future Work}
In this paper, 
we proposed the generative cooperative Net (GCN)  model which can generate unseen images with desired high-level visual concepts. %The GCN's classifier is here to facilitate the other part, the generator on high-level concept learning. 
Unlike the GANs, the GCN model does not have adversarial modules: the generator and classifier work cooperatively 
to minimize the objective function. 
Besides, the GCN model can used for data augmentation. Since the synthesized images are not simply the linear combination of the original images, it could provide an unique transformation on the target concept and preserve the concepts we hope to keep. In our case study of using the KDEF data, it transforms the person identity but keep the emotion. 
The GCN model looks like an alternative structure to the GANs, but it could be actually incorporated in the GANs framework and performed solely on its generator's training phase. Our future work would focus on this integration and investigate a more difficult task such as using small dataset to learn and enable emotion transformation for any new faces. We will also test more datasets to investigate how to fuse high-level concepts in other problems but not only image generation. 

\bibliography{egbib}
\end{document}